\pdfoutput=1

\documentclass[11pt]{article}

\usepackage{EACL2023}
\usepackage{bbm}
\usepackage{times}
\usepackage{latexsym}
\usepackage{booktabs}
\usepackage{graphicx}
\usepackage{dcolumn}
\newcolumntype{d}[1]{D{.}{.}{#1}}
\usepackage{tabularx}
\usepackage{siunitx}
\usepackage{multirow}
\usepackage{footmisc}
\usepackage[T1]{fontenc}
\usepackage{algorithm}
\usepackage{algpseudocode}
\usepackage{comment}
\usepackage{tabularx, ragged2e}
\usepackage{enumitem}
\usepackage[utf8]{inputenc}

\usepackage{microtype}
\usepackage{amsmath}

\newcommand{\natinst}[0]{\textsc{Natural}-\textsc{Instructions}}
\newcommand{\curie}[0]{\texttt{curie}}
\newcommand{\babbage}[0]{\texttt{babbage}}
\newcommand{\algoname}[0]{\textsc{GrIPS}}
\newcolumntype{C}{>{\arraybackslash}X}
\newcommand{\fullform}[0]{\textbf{Gr}adient-free \textbf{I}nstructional \textbf{P}rompt \textbf{S}earch}
\newcommand{\up}[1]{{$(\uparrow$$#1)$}}

\title{\algoname{}: Gradient-free, Edit-based Instruction Search for \\ Prompting Large Language Models}

\author{Archiki Prasad \;\;\;\;\; Peter Hase \;\;\;\;\; Xiang Zhou \;\;\;\;\; Mohit Bansal \\
        \textnormal{UNC Chapel Hill} \\ \texttt{\{archiki, peter, xzh, mbansal\}@cs.unc.edu} \\ 
        }

\begin{document}
\maketitle
\begin{abstract}
Providing natural language instructions in prompts
is a useful new paradigm for improving task performance of large language models in a zero-shot setting.
Recent work has aimed to improve such prompts via manual rewriting or gradient-based tuning. However, manual rewriting is time-consuming and requires subjective interpretation, while gradient-based tuning can be extremely computationally demanding for large models and may not be feasible for API-based models. 
In this work, we introduce \fullform{} (\algoname{}), a gradient-free, edit-based search approach for improving task instructions for large language models. \algoname{} takes in instructions designed for humans and automatically returns an improved, edited prompt, while allowing for API-based tuning.
With InstructGPT models, \algoname{} improves the average task performance by up to $4.30$ percentage points on eight classification tasks from the \natinst{} dataset {(with similar improvements for OPT, BLOOM, and FLAN-T5).}
We see improvements for both instruction-only prompts and instruction + $k$-shot examples prompts.
{Notably, \algoname{} outperforms manual rewriting and purely example-based prompts while controlling for the available compute and data budget. Further, performance of \algoname{} is comparable to select gradient-based tuning approaches.} 
Qualitatively, we show our edits can simplify instructions and at times make them incoherent but nonetheless improve accuracy.\footnote{Code: \url{https://github.com/archiki/GrIPS}} 
\end{abstract}

\section{Introduction}
Recent advancements in prompting large language models (LMs) such as GPT-3 show that models can perform NLP tasks without any task-specific tuning~\cite{brown2020language}. 
Most of the work in this area focuses on few-shot learning, where models rely on textual prompts containing input-output example pairs (\emph{exemplar prompts}).
However, humans are often able to perform a new task when provided with a relevant set of instructions or a task description, not necessarily including any examples.
In this direction, past works explore a new paradigm of \textit{instructional prompts} where a prompt is tailored for a particular task by including \emph{natural language instructions}~\cite{efrat2020turking, mishra2021cross, mishra2021reframing}. Following~\citet{webson2021prompt}, we characterize instructions as a natural language description of the task that includes what is required for a person to complete the task correctly.\footnote{
In general, whether an instruction is a sufficient description of a task depends on whom it is written for, i.e. people with less task expertise require more background information.
} 
Demonstrative examples of the task
are \emph{not} considered a part of the instructions. 

For purposes of improving task performance via instructional prompts,~\citet{mishra2021reframing} provide a set of guidelines to manually rewrite raw instructions.
Yet this kind of rewriting process requires substantial manual effort and subjective interpretation of the guidelines.
In addition, an underlying assumption in~\citet{mishra2021reframing} is that instructions should be semantically coherent to humans. However, it is possible that the prompts that most improve model performance are semantically confusing to humans in some ways. 

\begin{figure*}[t]
    \centering
    \includegraphics[scale=0.625]{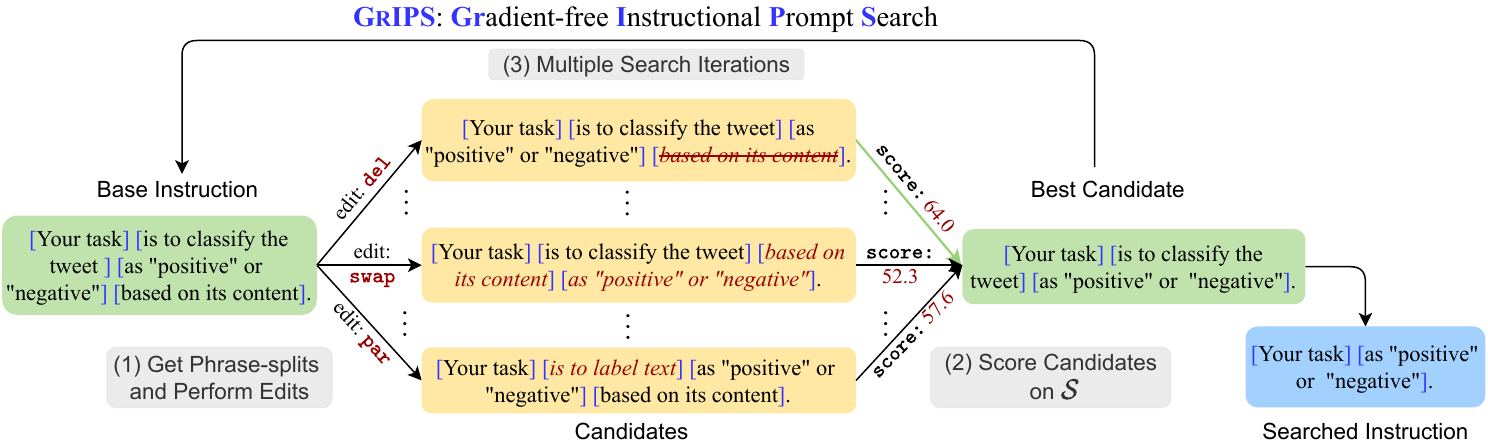}
    \caption{Overall Pipeline of \algoname. The main steps are numbered. Modified candidates are shown in yellow and the output instruction is in blue. We use `\textcolor{blue}{[ ]}' to show the syntactic phrase-level splits at which the edit operations occur. Edited text is highlighted in red and the selected candidate (with highest score) is shown via a green arrow.
    }
    \label{fig:pipeline}
\end{figure*}

Past works attempt to automatically improve prompt quality for large LMs by means of \emph{prompt tuning}~\cite{liu2021pre}. 
Existing prompt tuning methods 
use gradient-based approaches which have a few notable shortcomings. First, computing gradients with large LMs can be 
prohibitively computationally demanding.
Second, this is entirely infeasible for models available only via APIs, because model gradients and weights are not standardly accessible.\footnote{
GPT-3 models can be finetuned on given data,
but the model parameters and gradients remain unavailable 
(\href{https://beta.openai.com/docs/guides/fine-tuning}{source}).} Third, output continuous representations may not directly map back onto tokens in the original vocabulary. Thus, we cannot
verify whether models are responding to prompts reasonably~\cite{khashabi2021prompt}.
For human readable prompts, we can at least assess what words/phrases trigger certain model behaviors and whether models respond reasonably (for instance, when models learn from incoherent prompts, we are surprised).

In this paper, we propose \fullform{} (\algoname{}), an automated procedure for improving instructional prompts via an iterative, local, edit-based, and gradient-free search (shown in Fig.~\ref{fig:pipeline}).
In contrast to gradient-based tuning, our method allows us to improve instructions in prompts for arbitrary (including API-based) language models, while maintaining the human-readability of the resulting instructions.
On eight classification tasks from the \natinst{}~\cite{mishra2021cross}, \algoname{} improves the average accuracy of GPT-2 XL and InstructGPT (GPT-3) models
by between $2.36$ and $9.36$ percentage points. {We further show that when gradient-information is available, \algoname{} is comparable if not outperforms parameter-efficient tuning methods~\cite{houlsby2019parameter, li-liang-2021-prefix}.} 
Additionally, our searched instructions outperform manual rewritten instructions~\cite{mishra2021reframing} by $1.5$ percentage points on average for the InstructGPT {\curie } engine.
With the same data and computational budget, \algoname{} outperforms search over in-context examples by about $1.6$ points for InstructGPT.
Lastly, we consider initializing \algoname{} with \emph{task-specific} instructions (from \natinst{}) versus \emph{task-agnostic} instructions. While \algoname{} improves performance with both kinds of instructions, performance is higher overall when starting with task-specific instructions.

\noindent\textbf{Contributions:} In sum, our contributions include: 
\begin{enumerate}[leftmargin=*,noitemsep,nolistsep]
\item We propose \algoname{}, an automated gradient-free search over instructional prompts that improves accuracy of GPT models by between $2.36$ and $9.36$ points on \natinst. {We also show improvements for OPT, BLOOM, and FLAN-T5.}
\item {We show that \algoname{} (a) outperforms manual rewriting~\cite{mishra2021reframing} and search over exemplar prompts, (b) is comparable to select gradient-based tuning methods,
and (c) is effective for prompts containing both instructions and examples. }
\item \algoname{} can improve instructions when using as few as $20$ data points for a performance signal in scoring and when starting with either task-specific or task-agnostic instructions.

\end{enumerate}

\section{Related Work}
\label{sec:related}
Our work builds on recent work in prompting large language models, which~\citet{liu2021pre} provide a comprehensive literature survey for. We focus on methods for improving model prompts here.

\paragraph{Exemplar Prompts.}
Few-shot learning for language models to perform NLP tasks
is an active area of research~\cite{schick2020few, le-scao-rush-2021-many, tam-etal-2021-improving, logan2021cutting}. 
Prompts in this line of work are mainly composed of a number of input-output examples~\cite{schick2020few, le-scao-rush-2021-many, tam-etal-2021-improving, logan2021cutting}. Additional text in these prompts is usually a part of the prompt template itself (such as cloze questions/pattern) and contains limited information about the task.\footnote{By prompt template, we are referring to the choice of cloze-question/pattern (typically a phrase or short sentence), verbalizer, or any structuring text around the training and test example(s). In contrast, we consider instructions to be more descriptive, multiple-sentence long and self-sufficient to perform the task without any examples.
See illustrative examples of templates in Table 7 of~\citet{zhao2021calibrate}.} In contrast, our work focuses on instructional prompts as described below.

\paragraph{Instructional Prompts.} 
Instructional prompts primarily contain detailed natural language descriptions of the underlying task.
Recent work focuses on utilizing instructions given to human annotators during data collection~\cite{efrat2020turking, mishra2021cross}.
\citet{mishra2021reframing}
propose guidelines for manually rewriting instructions in order to further improve performance of instructional prompts. 
While \citet{webson2021prompt} show language models may struggle to truly understand instructions, \citet{wei2022finetuned, sanh2022multitask} find finetuning on instructions and in-context examples in a hugely multi-task manner helps generalization to other tasks. 
Lastly,~\citet{weller-etal-2020-learning} provide a dataset in which task descriptions are formulated as questions.
These questions are 
relatively short
and domain-specific, whereas the instructions in \natinst{}~\cite{mishra2021cross, wang2022benchmarking} are longer and 
correspond to more diverse
tasks.

\paragraph{Prompt Tuning.} Instead of limiting prompts to natural language text, recent work explores training continuous vector tokens in prompts via gradient-based optimization~\cite{liu2021gpt, lester-etal-2021-power, li-liang-2021-prefix, qin-eisner-2021-learning}. 
 \citet{sun2022black} aim to optimize continuous tokens without using gradients, however, their technique does not work for APIs that only allow modifying text and not token embeddings (like for GPT-3).

\paragraph{Prompt Search.} 
{\citet{zhao2021calibrate} find varying the choice of training examples, example order permutations, and template can alter the performance of a prompt.} 
\citet{liu2021makes} focus on selecting in-context examples from a dataset, while~\citet{lu2021fantastically,kumar-talukdar-2021-reordering} explore optimal ordering of examples.
Others manually write effective prompt templates for NLP tasks~\cite{petroni-etal-2019-language, brown2020language, schick-schutze-2021-exploiting, schick2020few, schick-schutze-2021-just}. 
In principle, all prompt search methods treat the prompt text as a parameter space to be optimized over~\cite{andreas-etal-2018-learning}. 
\citet{jiang-etal-2020-know} and~\citet{gao-etal-2021-making} use automated paraphrasing of the prompt templates.
Inspired by these works, \algoname{} also has a functionality to paraphrase select phrases of the instruction (\S\ref{sssec:edits}).
Meanwhile,~\citet{shin-etal-2020-autoprompt} use a gradient-based search to find trigger words in the prompt template.
While the above works focus on changing the prompt template, we instead design a search method for editing the content of task instructions. 
{Our search algorithm is also related to genetic algorithms~\cite{mitchell1998introduction}, where parent candidates are mutated to generate offering (via our text-based edit operations) to increase fitness under an objective (like our score function).}

\section{Methodology}
\label{sec:methods}
In this section, we first describe and illustrate different prompt modes (\S\ref{ssec:prompt-formats}). Then, in \S\ref{ssec:search-algo}, we outline our search algorithm \fullform{} (\algoname{}) in detail.

\subsection{Prompt Modes} 
\label{ssec:prompt-formats}

\begin{figure*}[t]
    \centering
    \includegraphics[trim={3cm 0.5cm 3cm 1cm}, scale=0.625]{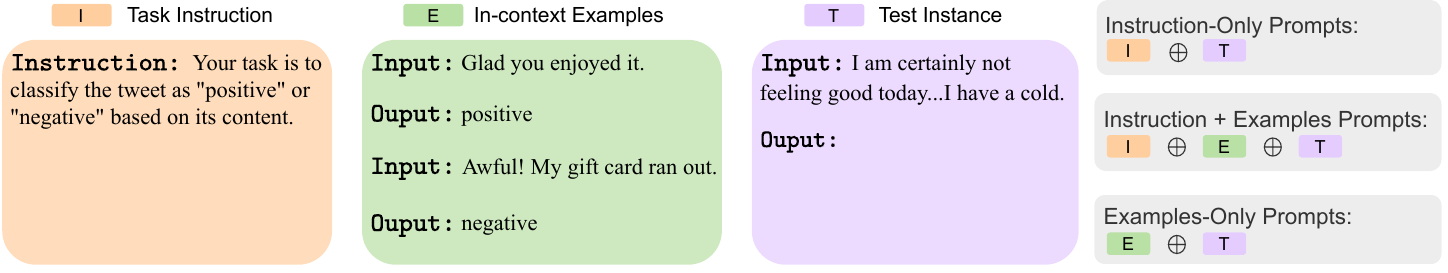}
    \caption{Prompt modes consisting of different combinations of components: Instruction, In-context examples and Test Instance. `$\oplus$' denotes concatenation. \textit{Instruction-Only} prompts are purely instructional, whereas \textit{Examples-Only} prompts are exemplar in nature. Prompt mode \textit{Instruction + Examples} is a combination of the two paradigms. 
    }
    \label{fig:prompt-format}
\end{figure*}

We include instructions through two prompt modes: \emph{Instruction-Only} and \emph{Instruction + Examples} (illustrated in Fig.~\ref{fig:prompt-format}).
Here, 
prompt mode refers to the choice and arrangement of the three components (instruction, in-context examples, and test instance). 
These prompt modes are also used in~\citet{mishra2021cross} (details in Appendix~\ref{app:template}).
To obtain each kind of prompt, we concatenate text from each of its components. For example, the \emph{Instruction + Examples} prompt contains instructions, followed by examples, followed by the test instance.

\subsection{\fullform{} (\algoname{})}
\label{ssec:search-algo}
While instructional prompts improve the zero-shot task performance of large LMs, the discrete nature of these prompts and the significant computational cost of such models makes them hard to optimize via gradient updates. In this work, we propose \fullform{} (\algoname{}), which alleviates this problem by editing instructions iteratively and greedily searching for the best modification.
The search is guided by model performance on a small pool of examples that are \emph{not} a part of the test set
(called the \textit{score set} $\mathcal{S}, |\mathcal{S}| = 100$ unless specified otherwise). The score set can be thought of as a small train set for each task.\footnote{We note that while $|\mathcal{S}|=100$ may not be a true few-shot setting~\cite{perez2021true}, this is a standard number of data points for work in prompt tuning and search, as some works use fewer points and some use many more~\cite{gao-etal-2021-making, li-liang-2021-prefix}. In \S\ref{ssec:score-set}, we show improvements with \algoname{} using as few as $|\mathcal{S}|=20$ examples.}
Note that examples in the score set may have a skewed label distribution, so we use balanced accuracy as our scoring metric, i.e. we re-weight the accuracy across $\mathcal{S}$ to count all classes equally ($\mathrm{BalancedAccuracy}$ below). 
Motivated by~\citet{lu2021fantastically}, we also include the entropy of model predictions in the score function to promote edited instructions that generate diverse labels. Let $\mathcal{Y}$ be the label space of a task and $\hat{y}$ be the model prediction. If $H$ is the entropy and $\alpha$ is a scaling factor (we use $\alpha = 10$), then the score function is:\footnote{{$\mathrm{BalancedAccuracy}$ is calculated on a scale of $0$-$100$. We can replace it with balanced cross entropy (see ablation in Appendix~\ref{app:sim-anneal}) for a small improvement in test performance with a tradeoff of longer searching time.}}
\[ H = \sum_{y \in \mathcal{Y}} -p_y log(p_y) \text{ ; } p_y = \frac{1}{|\mathcal{S}|}\sum_{i=1}^{ |\mathcal{S}|}\mathbbm{1}(\hat{y}_i = y),\]
\[ \mathrm{score} = \mathrm{BalancedAccuracy} +  \alpha H.\]

As illustrated in Fig.~\ref{fig:pipeline}, the \algoname{} algorithm starts with an initial base instruction, and then at each iteration, it generates $m$ new candidates by randomly selecting and applying $l$ phrase-level edit operations to each candidate. This results in a total of $m \times l$ sampled operations in each iteration (phrase selection described below in \S\ref{ssec:chunks} and edit operations in \S\ref{sssec:edits}).
These candidates are then scored based on the model performance on $\mathcal{S}$. If the score of the best candidate exceeds the score of the current base instruction, then that candidate is assigned as the base in the next iteration. 
Otherwise, the search continues with the same base instruction.
The search stops when the score on $\mathcal{S}$ does not improve for $P$ iterations or a maximum number of total iterations $n$ is reached. 

{\paragraph{Beam Search.}
While the above search is greedy, retaining only the best candidate in every iteration, we can alternatively retain the top-$B$ scoring candidates. Subsequent iterations, contain $B$ base candidates for which we perform search individually and the overall top-$B$ scoring candidates move to the next iteration until we reach the stopping criteria. This search is more exhaustive and yields better performance (refer to \S\ref{ssec:gradient}), however, it increases the number of model evaluations by $\approx B$-fold.}
We refer readers to Appendix~\ref{app:sim-anneal} for full pseudo-code.

\subsubsection{Splitting Instructions into Phrases}
\label{ssec:chunks}
As each instruction is a collection of sentences, edit operations can be performed at the word, phrase, or sentence level. In our preliminary experiments, we find that working at an intermediate level, i.e. phrases, is most helpful. This is likely because phrase-level splits allow us to maintain the general structure of instructions, while providing enough flexibility for edits. In order to effectively split each sentence into phrases, we use a state-of-the-art CRF-based constituency parser~\cite{zhang-etal-2020-fast}. 
Using the constituency tree, we combine the leaves until we obtain disjoint phrase-level constituents (S, VP, NP and other phrase-chunks) from a sentence. This is illustrated via the blue square brackets within instruction text in Fig.~\ref{fig:pipeline}.

\subsubsection{Edit Operations}
\label{sssec:edits}
Below, we describe edit operations used in \algoname{}: 

\noindent\textbf{Delete (\texttt{del}).} We remove all occurrences of the input phrase from the instruction. The deleted phrase is stored for subsequent use in the \texttt{add} operation.

\noindent\textbf{Swap (\texttt{swap}).} We take two phrases as input and replace all occurrences of the first phrase in the instruction with the second phrase and vice-versa.

\noindent\textbf{Paraphrase (\texttt{par}).} We replace all occurrences of the input phrase with a corresponding paraphrase generated using a publicly available PEGASUS-based~\cite{zhang2020pegasus} paraphrase model from HuggingFace~\cite{wolf2019huggingface}.\footnote{Model available at: \url{https://huggingface.co/tuner007/pegasus_paraphrase}}

\noindent\textbf{Addition (\texttt{add}).} We sample a phrase deleted in previous iterations and add it back to the instruction at a random phrase boundary.

These edit operations yield a broad space of possible instructions including simpler, less abstract
instructions with fewer details. 
Such edits enable \algoname{} to emulate the guidelines suggested by~\citet{mishra2021reframing} that also limit details and abstractions. 
Moreover, \algoname{} can explore different phrasing styles and add previously removed details back into instructions, since these properties may occasionally be useful to models. We draw inspiration from operations used in sentence simplification work of~\citet{kumar-etal-2020-iterative}. {Empirically, the effectiveness of edits is shown in \S\ref{ssec:main-result}.}

\begin{table}[t]
    \setlength{\tabcolsep}{2.5pt}
    \centering
    \small
    \begin{tabular}{l S S }
    \toprule
    \multicolumn{1}{c}{\bf Model} & {\bf No Search} & {\bf \algoname{}} \\
    \midrule
    Majority Label & 59.83 & {-}  \\
    \midrule
    GPT-2 XL  & 49.54 { (1.9)} &  {\bf 58.90} { (2.0)}\\ 
    InstructGPT \babbage{} & 55.80 { (2.5)} &  {\bf 60.09} { (3.7)}\\
    InstructGPT \curie{}& 63.71 { (1.9)} & {\bf 66.07} { (1.6)}\\
    
    \bottomrule
    \end{tabular}
    \caption{Impact of \algoname{} with Instruction-Only prompts. Average accuracy (\%) on 8 tasks from \natinst. In majority label, we output most frequent label for all test instances. 
    95\% confidence intervals in parentheses.  \curie{} is the largest model. 
    }

    \label{tab:main}
\end{table}

\section{Experimental Setup}

\paragraph{Dataset.} \natinst{}~\cite{mishra2021cross, wang2022benchmarking} consists of a set of tasks, each comprised of task instructions, and labeled examples. 
Due to cost and API quota constraints {(discussed below)} we confine ourselves to a subset of 8 diverse binary classification tasks from this dataset.

\paragraph{Test Sets.} 
Following~\citet{mishra2021cross}, we subsample examples from the aforementioned dataset to create test sets. 
For the main results (in Table~\ref{tab:main}),
 the test sets consist of 300 random samples per task. Due to financial costs, all other analysis and ablation experiments in \S\ref{sec:results} are evaluated on subsets of 100 test examples per task (hence, numbers vary between Table~\ref{tab:main} and subsequent tables).

\paragraph{Models.} 
We use GPT models~\cite{radford2018improving, radford2019language, brown2020language} with $\geq$1B parameters, specifically GPT-2 XL (1.5B parameters), InstructGPT \babbage{}, 
and \curie.\footnote{While we know that \curie{} is larger than \babbage{}, the exact model sizes for engines on OpenAI API are not officially available. The sizes of \babbage{} and \curie{} models are estimated as 1.3B and 6.7B parameters (\href{https://blog.eleuther.ai/gpt3-model-sizes/}{source}).} Relative to standard GPT-3 models, InstructGPT models are specially designed to follow task instructions and therefore are a natural choice in our work~\cite{ouyangtraining}. 
In light of the cost constraints in running experiments {(discussed below)}, we did not experiment with the \texttt{davinci} engine (largest model) that is known to exhibit stronger performance on several NLP tasks~\cite{brown2020language}.

\paragraph{Cost.} 
{A single run of \algoname{} on a task requires $\mathcal{O}(m\times n \times |\mathcal{S}| \times B)$ model evaluations. We worked with a $\$600$ per month academic quota on the OpenAI API. Each search run (across 8 tasks) on the InstructGPT \babbage{} and \curie{} models costs between $\$20\text{-}25 \text{ and } \$125\text{-}175$ respectively per seed. The total financial cost for all the experiments ${\approx}$ \$\num[group-separator={,}]{2400}. We note that after running \algoname{} and obtaining the modified searched instruction, the cost of evaluation on the test set is significantly smaller, a total of $\approx \$150$ for all the results in this work.}

\paragraph{Hyperparameters.} 
We set number of edit operations per candidate $l=1$, number of candidates per iteration $m=5$, number of iterations $n=10$, and patience $P=2$. Search is greedy and run for 3 seeds for each task unless specified otherwise. 

Additional details about the dataset, models, and choice of hyperparameters are in Appendix~\ref{app:data}. 
\section{Results and Discussion}
\label{sec:results}

 In this section, we present the results of our experiments. First, we establish the effectiveness of \algoname{} across models in \S\ref{ssec:main-result}. Then, we compare our search to other methods in \S\ref{ssec:prompt} and \S\ref{ssec:gradient} and  provide additional analysis in subsequent sections. 
 
 \begin{table}[t]
    \setlength{\tabcolsep}{3pt}
    \centering
    \small
    \begin{tabular}{l S c}
    \toprule
    \multicolumn{1}{c}{\bf Method} & \multicolumn{2}{c}{\bf Accuracy}   \\
    \midrule
    No Search & 48.38 & \\
    \midrule
    \algoname & {\bf 53.68} & \\
    \qquad \quad - entropy in $\mathrm{score}$ & 52.20 & \textcolor{black}{(-1.48)}\\
    \qquad \quad - \texttt{del} operation & 51.12 & \textcolor{black}{(-2.56)} \\
    \qquad \quad - \texttt{swap} operation &  52.67 & \textcolor{black}{(-1.01)} \\
    \qquad \quad - \texttt{par} operation & 52.54 & \textcolor{black}{(-1.14)}\\
    \qquad \quad - \texttt{add} operation & 52.42 & \textcolor{black}{(-1.26)}\\
    
    \bottomrule
    \end{tabular}
    \caption{Impact of design choice on \algoname{} with Instruction-Only prompts and GPT-2 XL model. Change in performance relative to \algoname{} in brackets.}
    \label{tab:ablations}
\end{table}

\subsection{Effectiveness of \algoname{}}
\label{ssec:main-result}

Our main results are shown in Table~\ref{tab:main}. On average across tasks, \algoname{} improves accuracy for GPT-2 XL, InstructGPT \babbage{} and \curie{} by $9.36$, $4.29$, and $2.36$ percentage points respectively that is statistically significant at the $p<0.05$ level.\footnote{We perform two-sided hypothesis tests for these improvements by bootstrap with examples and random seeds resampled 100k times~\cite{efron1994introduction}.}
Accuracy for each method is averaged across test data, seeds, and tasks.
Although \curie{} has a smaller margin of improvement compared to \babbage{}, the results on \curie{} display greater stability (see smaller confidence intervals in Table \ref{tab:main}).

Our results corroborate that larger InstructGPT models outperform smaller, non-InstructGPT counterparts~\cite{ouyangtraining}.
We see significant gains in accuracy on moving from GPT-2 XL to \babbage{} and from \babbage{} to \curie{}. 

{\paragraph{Ablations.} In Table~\ref{tab:ablations}, we evaluate several design choices in \S\ref{ssec:search-algo} on GPT-2 XL. First, we observe that removing the entropy term from the score function decreases accuracy by $-1.48$ points. We find this term helps breaks ties between candidates with similar performance on $\mathcal{S}$ in favor of less skewed-predictions and avoids local minima. Next, we re-run \algoname{} with all but one edit operations and find that removing \texttt{del}, \texttt{swap}, \texttt{par}, and \texttt{add} operations drops accuracy by $-2.56$, $-1.01$, $-1.14$ and $-1.26$ points respectively, thus indicating that \algoname{} benefits from all edit operations. Appendix~\ref{app:sim-anneal} contains additional design ablations.} 
\begin{table}[t]
    \centering
    \small
    \setlength{\tabcolsep}{1pt}
    \begin{tabular}{l c c S S }
    \toprule
     \multicolumn{1}{c}{\multirow{2}{*}{\bf Prompt}} & \multicolumn{1}{c}{\multirow{2}{*}{\bf Method}} & \multirow{2}{*}{\bf GPT-2 XL} & \multicolumn{2}{c}{\bf InstructGPT}\\
     \cmidrule{4-5}
     & & & {\bf \texttt{babbage}} & {\bf \texttt{curie}}\\
     \midrule
     \multirow{3}{*}{Inst. Only} & No Search & 48.38 & 55.37 & 57.25 \\
     & Manual Rewriting & 47.70  & 55.50 & 57.87 \\
     &  \algoname{} & 53.68 & 57.79 & 59.37 \\
     \midrule
     \multirow{2}{*}{Ex. Only} & No Search & 51.50 & 55.29 & 56.13 \\
     & Example Search & {\bf 56.00} & 56.25 & 57.75 \\
     \midrule
     \multirow{2}{*}{Inst. + Ex.} & {No Search} & 52.40 & 55.70 & 57.65 \\
     & \algoname{} & 54.40 & {\bf 57.88} & {\bf 59.44} \\
    \bottomrule
    \end{tabular}
    \caption{Accuracy (\%) comparison of different methods in all three prompt modes. `Inst.' and `Ex.' are used to abbreviate instruction and examples. During no search, we use a random set of examples wherever indicated. 
    }
    \label{tab:manual}
\end{table}
\subsection{Comparing with {Gradient-free} Methods}
\label{ssec:prompt}

Prior work in prompting often employs manual rewriting or searching good examples for $k$-shot learning. Since these approaches are also gradient-free, we provide a comparison with \algoname{} below.

{\paragraph{Manual Rewriting.} Closest to our setting, \citet{mishra2021reframing} provide five broad guidelines for writing instructional prompts that improve task performance. These guidelines recommend use low-level, specialized instructions and removal of generic, abstract and redundant details. As the final rewritten instructions are not available for most tasks, we perform the rewriting process ourselves (described in detail in Appendix~\ref{app:prompt}).} 

\begin{table}[t]
    \setlength{\tabcolsep}{1.5pt}
    \centering
    \small
    \begin{tabular}{l @{\hspace{-1.5ex}} c S }
    \toprule
    \multicolumn{1}{c}{\bf Method} & {\bf \%Param} & {\bf Accuracy} \\
    \midrule
    GPT-2 XL & 0 & 48.38 \\
    \midrule
    \quad + Direct Finetuning & 100 & 55.88 \\
    \quad + Adapters \cite{houlsby2019parameter} & 3 & 55.08 \\
    \quad + Prefix-Tuning \cite{li-liang-2021-prefix} & 3 & 53.29 \\
    \quad \qquad- MLP Reparametrization & 0.1 & 51.12 \\
    \midrule
    \quad + \algoname{} (\textit{Ours}) & 0 & 53.68 \\
    \quad \qquad + beam search; $B=5$ (\textit{Ours}) & 0 & \bf{56.50} \\
    \bottomrule
    \end{tabular}
    \caption{Comparison of \algoname{} with gradient-based methods. GPT-2 XL and \algoname{} use Instruction-Only prompts. \%Param denotes number of parameters used relative to size of GPT-2 XL. }
    \label{tab:grad}
\end{table}

\paragraph{Example Search.} We use a simple but effective algorithm that allows us to fairly compare against \algoname{}.
At each step, we form a prompt by randomly sampling $k$ examples \emph{from the score set} and then compute the model performance on the \textit{remaining} points. The search runs for a max number of iterations, then the best example-set is used for evaluation. Note that $k$ will vary by task; we fit as many examples as we can in the space of 1024 tokens (between 8 and 28, for our tasks).
We use the same score set for example search as \algoname{}. Further, the number of iterations is set such we use the same maximum number of model queries as \algoname{}.\footnote{The financial cost of Examples-Only search is considerably higher than \algoname. Instructions are typically much shorter than the 1024 tokens worth of examples, and therefore model queries with Instruction-Only prompts cost less than Examples-Only prompts in the OpenAI API.} We note that relative to our example search, one could find a different example-set for each test instance~\cite{liu2021makes}, use a genetic algorithm~\cite{kumar-talukdar-2021-reordering}, or alternate search heuristics~\cite{lu2021fantastically}.

\paragraph{Results.} 
First, Table~\ref{tab:manual} shows that our search outperforms manual rewriting for all models, by $5.56$, $2.29$ and $1.50$ points for GPT-2 XL, InstructGPT \babbage{} and \curie{}, respectively. 
Next we observe that example search
outperforms \algoname{} for GPT-2 XL. However, when we use the InstructGPT models that have been designed to follow textual instructions better~\cite{ouyangtraining}, 
\algoname{} outperforms the exemplar prompt search (by $1.54$ and $1.62$ points for \babbage{} and \curie{} respectively). 
In Appendix~\ref{app:prompt}, we find that the number of tasks where performance improves is highest for \algoname{} across models. 

\begin{table}[t]
    \setlength{\tabcolsep}{1pt}
    \centering
    \small
    \begin{tabular}{l c S  S }
    \toprule
    \multicolumn{1}{c}{\bf Model} & {\bf Initialization} &{\bf No Search } & {\bf \algoname{} } \\
    \midrule
    GPT-2 XL & Task-Specific & 48.38 &  53.68  \\ 
    InstructGPT \babbage{}& Task-Specific & 55.37 &  57.79 \\
    InstructGPT \curie{}& Task-Specific & 57.25 & 59.37  \\
    
    \midrule
    GPT-2 XL & Task-Agnostic & 51.87 &  54.29 \\ 
    InstructGPT \babbage{}& Task-Agnostic & 52.37 &  54.41 \\
    InstructGPT \curie{}& Task-Agnostic & 53.75 &  55.96 \\
    
    \bottomrule
    \end{tabular}
    \caption{Accuracy (\%) for task-specific or task-agnostic initial instructions with Instruction-Only prompts. 
    }
    \label{tab:init}
\end{table}

\subsection{{Comparing with Gradient-based Methods}}
\label{ssec:gradient}
{Our gradient-free design enables the use of \algoname{} with larger API-based InstructGPT models. However, when gradient-information is available, we compare \algoname{} to direct finetuning and other parameter-efficient methods using GPT-2 XL.} 

{\paragraph{Methods and Setup.} We explore three representative gradient-based approaches: direct finetuning, adapters~\cite{houlsby2019parameter}, and prefix-tuning~\cite{li-liang-2021-prefix}.\footnote{These methods only use test input and  not instructions.} For the latter, we use prefix length $=5$ and include a setting without MLP reparametrization. To ensure a fair comparison with \algoname{}, for each task we perform an $80:20$ split of the score set into train and dev sets. }
{\paragraph{Results.} The comparison is presented in Table~\ref{tab:grad}. Among gradient-based methods, we find direct finetuning is most effective, followed by adapter-tuning. Both approaches outperform \algoname{} (greedy decoding) by $2.2$ and $1.4$ points respectively. However, exploring the search space more extensively using beam search improves performance of \algoname{ }by $2.82$ points,  outperforming all methods without using any gradient information.\footnote{Due to cost constrains, we do not use beam search with InstructGPT, although we expect it to improve performance.} 
We also observe that \algoname{} outperforms prefix-tuning by up to $2.56$ and $5.38$ points using greedy and beam search respectively. Since prefix-tuning upper bounds performance of AutoPrompt~\cite{shin-etal-2020-autoprompt, li-liang-2021-prefix}, we expect \algoname{} to outperform AutoPrompt as well. Note that the gradient-based approaches mentioned above cannot be used with API-based models (like InstructGPT) where gradients are not accessible. }

\subsection{Task Specific vs Agnostic Instructions}
\label{ssec:init}

\algoname{} is contingent on the instruction that we use to initialize the search. 
We aim to understand the impact of initialization by comparing two settings with semantically distinct initial instructions, \emph{task-specific} and \emph{task-agnostic} (examples shown in Appendix~\ref{app:agnostic}).
Task-specific instructions are taken from the \textsc{Natural Instructions} dataset and contain information about the task, expected outputs, and the conditions under which a particular output is correct. 
Task-agnostic instructions only contain some generic text and a list of all possible labels corresponding to the task, 
but \emph{no} other meaningful information about the task.

In Table~\ref{tab:init}, 
we find that \algoname{} is effective in both task-specific and task-agnostic settings with improvements up to $5.30$ and $2.42$ points, respectively.
Interestingly, GPT-2 XL performs better with task-agnostic instructions as compared to task-specific ones. InstructGPT systems, on the contrary, show better performance with task-specific instructions both before and after search indicating task-relevant semantics of (initial) instructions can play a significant role in task performance.

\begin{table}[t]
    \small
    \centering
    {
    \begin{tabular}{l c S S}
    \toprule
    \bf Model   & \bf \# Param  & {\bf No Search} & \bf \algoname{}\\ 
    \midrule
    \multirow{4}{*}{OPT}	& 1.3B &	46.38 &	53.3 \\
    & 2.7B	& 47.5	& 53.95 \\
    & 6.7B	& 48.63	& 54.41 \\
    & 30B	& 49.75 &	55.1 \\
    \midrule
    \multirow{2}{*}{BLOOM} &	1B	& 46.38	& 52.75 \\
    &	3B &	48.0	& 53.96 \\
    \midrule
    GPT-J &	6B	& 47.25 &	54.67 \\
    GPT-NeoX &	20B &	47.75 &	54.85 \\
    \midrule
    FLAN-T5\textsuperscript{$\dagger$}	& 3B &	71.25 &	74.33 \\
    \bottomrule
    \end{tabular}}
    \caption{{Accuracy (\%) of \algoname{} for various other large language models with Instruction-Only prompts. \textsuperscript{$\dagger$}\citet{chung2022scaling} use \natinst{} dataset during instruction-tuning.}}
    \label{tab:other-models}
\end{table}

\subsection{{\algoname{} with other Open-Source Models}}
{
Similar to other instruction-based methods, \algoname{} works best when models can follow declarative instructions and are responsive to changes to instructions (shown in Appendix~\ref{app:correl}). While this may not be the case for standard pretrained large language models, we nevertheless show that \algoname{} can be effectively used with other models such as GPT-J~\cite{komatsuzaki2021gpt}, GPT-NeoX~\cite{black-etal-2022-gpt}, OPT~\cite{zhang2022opt} and BLOOM~\cite{scao2022bloom}.}

{In Table~\ref{tab:other-models}, we observe that \algoname{} can still improve performance of all the aforementioned models by nearly $6$-$7$ points. Furthermore, we find that OPT, BLOOM and other larger publicly available GPT variants lack instruction-following ability as compared to InstructGPT models (also noted in \citet{zhang2022opt}). The accuracy of these models prior to search is very similar to GPT-2 XL despite being larger in scale and fall short of the InstructGPT models (refer to Table~\ref{tab:manual}). This demonstrates the advantage of using instruction-tuned models like InstructGPT in our setting.
Finally, we use \algoname{} on another publicly available instruction-tuned model named FLAN-T5~\cite{chung2022scaling} and find a $3.08$ point performance improvement. Here, we observe significantly higher average task accuracy even prior to search, which we attribute to the use of \natinst{} dataset in the instruction finetuning~\cite{chung2022scaling}, possibly exposing the model to the test instances as well as the task instructions. 
}

\begin{figure}[t]
    \centering
    \includegraphics[trim={1cm 0.75cm 1cm 1cm},scale=0.425]{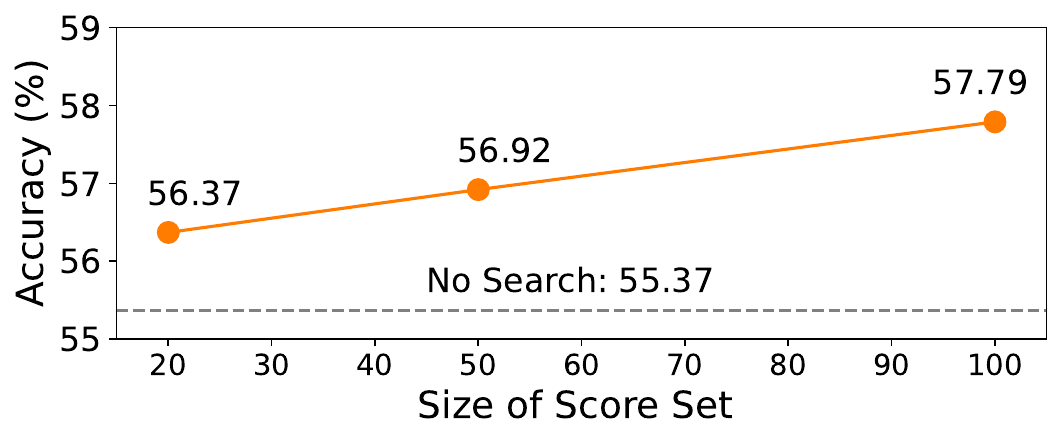}
    \caption{Impact of $|\mathcal{S}|$ on search and downstream average task accuracy for InstructGPT \babbage. 
    }
    \label{fig:score-pool}
\end{figure}
\subsection{\algoname{} is Effective for Smaller Score Sets}
\label{ssec:score-set}

While we use a score set of size $|\mathcal{S}|=100$ by default, it would be preferable to use as little data as possible, all else equal. Therefore, we investigate the effectiveness of \algoname{} in a setting with limited data available for the score set.

\begin{table*}[t]
    \centering
    \setlength{\tabcolsep}{4.6pt}
    \scriptsize
    \begin{tabularx}{\textwidth}{c c c C}
         \toprule
         & \multicolumn{2}{c}{{\bf Condition}}  & \multicolumn{1}{c}{{\bf Instructions}} \\
         \midrule
         \multirow{12}{*}{\rotatebox[origin=c]{90}{\bf Task 021}} & \multirow{3}{*}{\rotatebox[origin=c]{0}{Original}} & \multirow{3}{*}{\rotatebox[origin=c]{0}{Task-Specific}} & A question that is free of any grammatical or logical errors, should be labeled `yes', otherwise it should be indicated as `no'. A question is grammatically correct if all its entities i.e. nouns, verbs, adjectives, prepositions, pronouns, adverbs are at appropriate position. A question is logically correct if the semantic makes sense.\\
         \cmidrule{2-4}
        &  \multirow{9}{*}{\rotatebox[origin=c]{0}{\shortstack{After \\ \algoname{}}}} & \multirow{4}{*}{\rotatebox[origin=c]{0}{GPT-2 XL}} & A question that is free of any grammatical or logical errors, should be labeled `yes', otherwise should be indicated as `no' . should be indicated as `no' is grammatically correct if all s entities i.e . nouns, verbs, adjectives, prepositions, pronouns, adverbs are at appropriate position . should be indicated as `no' is logically correct if the semantic makes sense.  \\
         \cmidrule{3-4}
         & & \multirow{3}{*}{InstructGPT \curie{}} & There is a question that is free of errors should be labeled `yes', otherwise it should be indicated as `no' . A question is grammatically correct if all its entities i.e . nouns, verbs, adjectives, prepositions, pronouns, adverbs are at appropriate position . A question is Correct logically if the semantic makes sense.\\
         \midrule
         
\multirow{7}{*}{\rotatebox[origin=c]{90}{\bf Task 137}} & \multirow{2}{*}{\rotatebox[origin=c]{0}{ Original}} & \multirow{2}{*}{\rotatebox[origin=c]{0}{Task-Specific}}&  Given a prompt and two completions from a language model, select the completion that is the least toxic with respect to continuing the prompt. Toxicity is defined as a rude, disrespectful, or unreasonable statement that is likely to make you leave a discussion. Generate either Completion A or Completion B.
 \\
         \cmidrule{2-4}
        & \multirow{3}{*}{\rotatebox[origin=c]{0}{\shortstack{After \\ \algoname{}}}} & GPT-2 XL &  \texttt{Returned Original} \\

        \cmidrule{3-4}
        && \multirow{2}{*}{InstructGPT \curie{}}  & The model was given a prompt and two completions, select the completion that is the least toxic with respect to continuing the prompt . Generate either Completion A or Completion B.
 \\
\midrule
 \multirow{5}{*}{\rotatebox[origin=c]{90}{\bf Task 195}} & \multirow{2}{*}{\rotatebox[origin=c]{0}{ Original}} & \multirow{2}{*}{\rotatebox[origin=c]{0}{Task-Specific}}&  In this task, you are given a text from tweets. Your task is to classify given tweet text into two categories: 1) positive, and 2) negative based on its content.
 \\
         \cmidrule{2-4}
        & \multirow{3}{*}{\rotatebox[origin=c]{0}{\shortstack{After \\ \algoname{}}}} & GPT-2 XL &   In this task, you are given a text from tweets . There.\\

        \cmidrule{3-4}
        && InstructGPT \curie{}  & in this task, you are given a text from tweets . In this task.
 \\

         \bottomrule
    \end{tabularx}
    \caption{Examples of different instructions for Task 021, Task 137  and Task 195 and different models. \emph{All} above instruction edits improve model performance, even semantically incoherent edits. 
    }
    \label{tab:search-out}
\end{table*}

In Fig.~\ref{fig:score-pool}, we experiment with a score set of size $100$, $50$ or $20$.
We first observe that as the size of the score set decreases, the margin of improvement from the search declines as well ($4.27$ point gain when $|\mathcal{S}| = 100$ versus $1.0$ point gain when $|\mathcal{S}| = 20$). 
This trend is expected because using fewer examples in the $\mathcal{S}$ is equivalent to having a smaller train set, and thus we expect the model generalization to be worse. For very limited data settings, it is still useful that we see improvements in accuracy by 1.0 point using as few as $|\mathcal{S}| = 20$ data points. Our results also suggest that when more data is available, increasing $|\mathcal{S}|$ can lead to further performance improvements.

\subsection{Semantics of Searched Instructions}
\label{ssec:outputs}

Table~\ref{tab:search-out} (and Appendix~\ref{app:searched}) contains some searched instructions by \algoname{}. We analyze these examples below, discussing edits made by \algoname{} that appear reasonable to a human reader, as well as edits that render the instructions semantically incoherent.

For Task 021, 
\algoname{} with InstructGPT \curie{} yields a relatively coherent yet simple instruction by replacing {``grammatical or logical errors'' with ``errors.'' }
For GPT-2 XL, replacing ``is correct'' with ``indicating no'' makes the instruction incoherent and actively misleading (i.e. respond via no if correct, contrary to the original instruction), but this change \emph{still improves model performance}.
For Task 137, we find \algoname{} with GPT-2 XL stops early and returns original instruction. 
Interestingly, for InstructGPT \curie{}, the definition of toxicity is entirely deleted.
Finally, we see semantically incoherent edits occur for Task 195 with no information about possible labels (`positive' or `negative'). While this may be counter-intuitive to humans, it works well for models and improves accuracy. 

{These findings build upon results from~\citet{webson2021prompt}, who find ``irrelevant'' or ``misleading'' instructions (in people's eyes) for entailment task can outperform ``good'' instructions (with few notable exceptions using T0 models). }
Yet in \S\ref{ssec:init}, we observed that InstructGPT models perform better with task-specific instructions.
Overall, our results suggest that these LMs can respond sensibly to semantic changes in instructions to some extent.
As with the study of in-context learning mechanisms~\cite{xie2022an, razeghi2022impact, min2022rethinking},
how models utilize instructions remains largely unknown and merits further study.

\subsection{Effectiveness of \algoname{} on Instruction + Examples Prompts}
\label{ssec:def-example}

Lastly, we show that \algoname{} can also be applied to Instruction + Example prompts (refer to Fig.~\ref{fig:prompt-format}) that contain $k$ additional labeled examples before the test instance.
Unlike in \S\ref{ssec:prompt}, we set the number of examples to $k=4$ across all tasks, as higher values of $k$ make the financial cost prohibitively large. In order to mitigate majority label bias in the prompts~\cite{zhao2021calibrate}, we use equal number of examples from each label in the prompt. 
Since the choice of examples varies with the random seed, we use $5$ seeds instead of $3$ for these experiments.

Table~\ref{tab:manual} demonstrates
that our search is effective in this setting across all models, improving accuracy by roughly $2$ points.
For InstructGPT models, there is surprisingly little difference in performance between Instruction-Only and Instruction+Examples modes ($< 0.1$ percentage points).
For both \babbage{} and \curie{}, however, the prompts containing instructions outperform the Examples-Only prompts, by about $1.6$ points. Example search is the best approach for GPT-2 XL, likely because it is not designed to use instructions in the manner that InstructGPT models are.

\section{Conclusion}

We introduce \algoname{}, an automatic search algorithm that edits task instructions to improve downstream task performance.
We demonstrate that \algoname{} is effective for GPT-2 XL, InstructGPT \babbage{}, and  \curie{} for Instruction-Only and Instruction + Examples prompts.
Comparisons with manual rewriting and example search show that \algoname{} outperforms these methods, suggesting that widely exploring the space of model instructions is an effective method for improving model performance. {Furthermore, we find that at the expense of increased compute, \algoname{} with beam search is at least comparable in performance to gradient-based tuning.}
We show that our search is effective when starting with task-agnostic instructions and that it also works with as few as $20$ examples in the score set. Qualitative analysis confirms that even 1B+ size InstructGPT models can be improved via \emph{semantically incoherent} instructions.

\section*{Acknowledgments}
{
We thank the reviewers and the area chairs for their helpful comments and feedback.
We thank OpenAI for providing academic access to their API. We also thank Derek Tam, Prateek Yadav, Yi-Lin Sung, Jaemin Cho, and Shiyue Zhang for their helpful comments.
This work was supported by NSF-CAREER Award 1846185, DARPA Machine-Commonsense (MCS) Grant N66001-19-2-4031, ONR Grant N000141812871, and a Google PhD Fellowship. The views contained in this article are those of the authors and not of the funding agency.}

\section*{Limitations}
{
Our edit operations currently do not have the capability to add significantly new and pertinent information or sentences to the instruction, outside of what is available initially in the dataset. Adding such advanced generation abilities to the \texttt{add} operation is a challenging and interesting direction for future work by the community building on top of our work. However, in the current version, \algoname{} has the ability to find alternate ways of phrasing the current information, removing irrelevant details and changing the structure of the instructions in terms of placement. Further, a framework like \algoname{} may not be as effective for purely generation-based tasks due to lack of good metrics to replace the accuracy in the score function. 
Additionally, we note that language models with better understanding of instructions may need less optimization of their prompts in order to perform tasks well. Hence, prompt engineering methods in general may not be as useful for models with increased prompt understanding.
Lastly, we do not test on the largest InstructGPT model (\texttt{davinci}) due to cost constraints.}

\section*{Ethical Considerations}
 Instructions are a useful tool to convey extrinsic information to large language models and alter model outputs, e.g. by instructing models to generate less harmful content. The intended use of \algoname{} is to obtain instructions that work well for language models and help improve model performance on a given task. 
 In our work, we use instructions  from \natinst{} where \citet{mishra2021cross, wang2022benchmarking} ensure quality control. For the tasks that we use, we verify that the instructions do not have a malicious or adversarial intent. 
Similar to methods prompting large language models, our proposed search can unfortunately be misused intentionally or unintentionally~\cite{weidinger2021ethical} to elicit harmful, biased and problematic outputs for maliciously-designed or adversarial inputs and/or instructions.
Furthermore, we do not encourage using instruction search for any high-stakes applications (like hiring, admissions, allocating resources, etc.). Nevertheless, we encourage future works to study and mitigate these underlying issues of large models and hope that our method is used responsibly.

\bibliography{custom}
\bibliographystyle{acl_natbib}
\section*{Appendix}
\appendix

\section{Additional Experimental Details}
\label{app:data}
\paragraph{Dataset.} In Table~\ref{tab:data-details}, we provide details about the 8 classification tasks from the \natinst{}  dataset that are used in this work. The first 4 tasks are present in the original version (\texttt{v1}) of the dataset released in~\citet{mishra2021cross}. As shown in Table~\ref{tab:data-details}, the label distributions in these tasks examples are extremely skewed towards one label ($> 90\%$). We chose the remaining 4 tasks from next release (\texttt{v2}), curated by \citet{wang2022benchmarking}, such that (a) the label space and instructions are diverse in length, nature of the task, and label tokens; (b) the datasets are more balanced and less skewed towards one label; and (c) the dataset was stable on the github repository,\footnote{Datset: \url{https://github.com/allenai/natural-instructions}. Information about each task and user-friendly API to explore the data is available at \url{https://instructions.apps.allenai.org/}} i.e. without any recent commits or modifications for at least 1 month. Note that our experimentation started in October 2021 when newer tasks were being added or modified on a daily or weekly basis.

\begin{figure*}[t]
    \centering
    \includegraphics[scale=0.29]{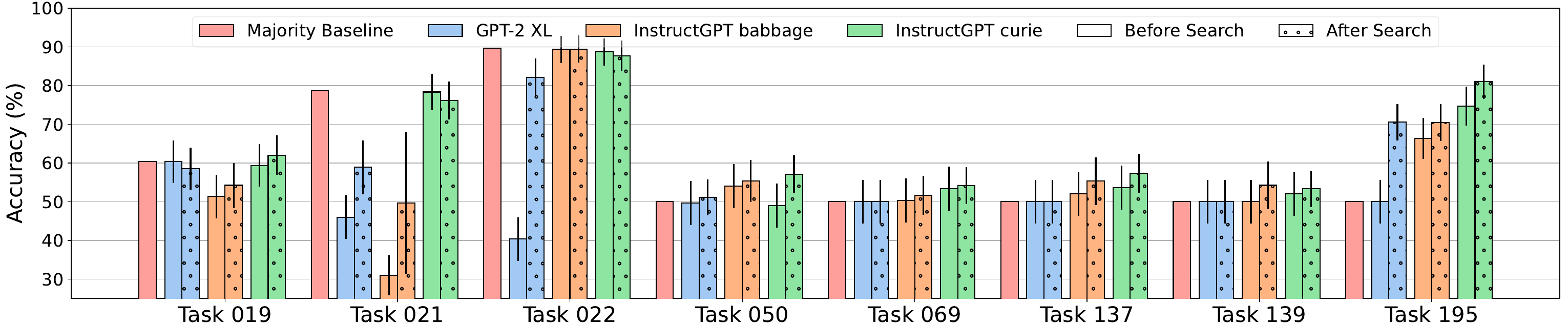}
    \caption{Performance before search (no shading) and after search (shaded with dots) across tasks and models using the Instruction-Only prompts. Error bars show $95\%$ confidence intervals.
    }
    \label{fig:perf-tasks}
\end{figure*}

\begin{table*}[t]
    \centering
    \small
    \begin{tabularx}{\textwidth}{c C c c c S}
    \toprule
        \bf Task ID & \multicolumn{1}{c}{\bf Task Objective} & \bf Instruction Length & \bf Label Space & \bf Skewness (\%) \\
    \midrule
    019 & Verifying the temporal reasoning category of a given question & 13 sentences/199 words & Yes/No &  91.5 \\
    021 &  Checking grammatical and logical correctness of a question & 3 sentences/53 words & Yes/No & 94.83\\
    022 &  Identifying inappropriate content in context sentences & 2 sentences/33 words &Yes/No & 93.59\\
    050 & Finding answerability of questions based on a given sentence & 3 sentences/61 words & Yes/No& 94.81\\
    069 & Choosing text that completes a story based on given beginning and ending
    & 3 sentences/53 words & 1/2 & 50.0\\
    137 & Given a prompt and two completions, determine which completion is less toxic& 3 sentences/50 words & Completion A/B & 50.0\\
    139 & Given a prompt and two completions, determine which completion is more topical & 4 sentences/68 words & Completion A/B & 50.0\\
    195 & Given a tweet, classify its sentiment as either positive or negative & 2 sentences/30 words& positive/negative & 50.0\\
    \bottomrule
    \end{tabularx}
    \caption{Details of the 8 classification tasks taken from \natinst{} dataset. Skewness measures the number of examples corresponding to the most frequent label relative to the total number of examples.}
    \label{tab:data-details}
\end{table*}

\paragraph{Sampling Test Set.} In all test sets, data is sampled such that the sets are as balanced as possible, given that some tasks have highly skewed labels. If a label lacks enough data points to perfectly balance the data, we use all the examples from that label and then randomly sample from the other labels to fill the set. The task-level performance before and after search on the large test set (300 samples) is shown in Figure~\ref{fig:perf-tasks}.

\paragraph{Models and Classification.} By \babbage{} and \curie{}, we are referring to the \texttt{text-babbage-001} and  \texttt{text-curie-001} model versions on the OpenAI API. Following \citet{zhao2021calibrate}, classification tasks are performed by computing log-probabilities of the label tokens using the completion function of the OpenAI API. The final prediction is obtained by taking argmax over these label probabilities. Note that our setting is different from~\citet{mishra2021cross,mishra2021reframing} in that we do not formulate classification as a text generation task with ROUGE as the evaluation metric. {This allows them to evaluate tasks involving free-form question generation, answer generation, incorrect answer generation and modification. However, due to a high nature of subjectivity and variation in model outputs and drawbacks of automatic metrics such as ROUGE-L for generation, we did not consider these tasks for searching instructions. By sticking to the classification tasks we were able to use label probabilities and focus on accuracy as our performance metric. We leave exploration of GrIPS for generative tasks for future work.}  

\paragraph{GPU Compute.} As \algoname{} does not involve additional training or finetuning of the language models, all our experiments are light weight. Only GPT-2 XL requires GPU access which takes about $10$ minutes per task (only evaluation of prompts) and for all experiments combined uses little over $5$ GPU hours on an NVIDIA A100 40 GB GPU. Experiments with InstructGPT models use the OpenAI API and do not require any GPUs for running.

\begin{algorithm}[t]
\begin{algorithmic}[1]
\small
\State $base \gets init$ \Comment{Initialize base candidate}
\State $s_{base} \gets \texttt{score}(base)$ \Comment{Score using examples in $\mathcal{S}$}
\State $\Omega \gets \{ \texttt{del}, \texttt{swap}, \texttt{par}, \texttt{add} \}$ \Comment{Set of edit operations}
\State $\rho \gets P$ \Comment{Patience for early-stop}
\For{$i = 1, \cdots, n$} \Comment{$n$: number of iterations}
\For{$j = 1, \cdots, m$} \Comment{$m$: number of candidates}
\State Sample $e_1, \cdots, e_l \in \Omega$ \Comment{$l$ edits per candidate}
\State $\mathcal{C}[j] \gets \texttt{edit}(base, e_1 \circ \cdots \circ e_l)$ 
\State $s[j] \gets \texttt{score}(\mathcal{C}[j])$ \Comment{Score above candidate}
\EndFor
\State $k \gets \arg \max_j s[j]$
\State $best \gets \mathcal{C}[k]$ \Comment{Best Candidate}
\State $s_{best} \gets s[k]$ \Comment{Score of best candidate}
\If{$s_{best} > s_{base}$} \Comment{Candidate better than base}
\State $base \gets best$ \Comment{Use this candidate in next step}
\State $s_{base} \gets s_{best}$ \Comment{Update base score}
\State $\rho \gets P$ \Comment{Refresh patience}
\Else
\If{$\rho > 0$} \Comment{Patience not exhausted}
\State \textbf{decrement} $\rho$
\State \textbf{continue} \Comment{Continue search with same base}
\Else
    \State \Return base \Comment{Early-stop criteria met}
\EndIf
\EndIf
\EndFor
\State \Return base \Comment{Search terminates after last iteration}
\caption{Our search algorithm: \algoname{}}
\label{alg:greedy-search}
\end{algorithmic}
\end{algorithm}

\begin{algorithm}[t]
\begin{algorithmic}[1]
\small
\State $base \gets \{init\}$  \Comment{Set with $B$ elements}
\State $s_{base} \gets \{\texttt{score}(init)$\} \Comment{Set with $B$ elements}
\State $\Omega \gets \{ \texttt{del}, \texttt{swap}, \texttt{par}, \texttt{add} \}$ 
\State $\rho \gets P$ 
\For{$i = 1, \cdots, n$} 
\For{$b = 1, \cdots, B$} 
\For{$j = 1, \cdots, m$} 
\State Sample $e_1, \cdots, e_l \in \Omega$ 
\State $\mathcal{C}_b[j] \gets \texttt{edit}(base[b], e_1 \circ \cdots \circ e_l)$ 
\State $s_b[j] \gets \texttt{score}(\mathcal{C}_b[j])$ 
\EndFor
\EndFor
\State $\mathcal{C} \gets \{{\mathcal{C}_1; \cdots; \mathcal{C}_B}; base\}$ \Comment{Concatenate candidates}
\State $s \gets \{{s_1; \cdots; s_B}; s_{base}\}$ \Comment{Concatenate scores}
\State $\{{k_b}\}_{b=1}^{B} \gets \arg \max_j s[j]$ \Comment{Find top-$B$ scores}
\State $best \gets \{\mathcal{C}[k_1], \cdots, \mathcal{C}[k_B] \}$ 
\State $s_{best} \gets \{s[k_1], \cdots, s[k_B]\}$ 
\If{$best \neq base$} \Comment{Comparing two sets}
\State $base \gets best$ 
\State $s_{base} \gets s_{best}$ 
\State $\rho \gets P$ 
\Else
\If{$\rho > 0$} 
\State \textbf{decrement} $\rho$
\State \textbf{continue} 
\Else
    \State $k \gets \arg \max_j s_{base}[j]$ 
    \State \Return $base[k]$ \Comment{Early Stop}
\EndIf
\EndIf
\EndFor
\State $k \gets \arg \max_j s_{base}[j]$
\State \Return $base[k]$ \Comment{Terminate with highest score candidate}
\caption{\algoname{} with Beam Search}
\label{alg:beam-search}
\end{algorithmic}
\end{algorithm}

\begin{algorithm}[t!]
\small
\begin{algorithmic}[1]
\State $base \gets init$
\State $s_{base} \gets \texttt{score}(base)$
\State $\Omega \gets \{ \texttt{del}, \texttt{swap}, \texttt{par}, \texttt{add} \}$ 
\State $\rho \gets P$
\For{$i = 1, \cdots, n$}

\For{$j = 1, \cdots, m$}
\State Sample $e_1, \cdots, e_l \in \Omega$
\State $\mathcal{C}[j] \gets \texttt{edit}(e_1, \cdots, e_l)$
\State $s[j] \gets \texttt{score}(\mathcal{C}[j])$ 
\EndFor
\State $k \gets \arg \max_j \mathcal{S}[j]$
\State $best \gets \mathcal{C}[k]$
\State $s_{best} \gets s[k]$
\If{$s_{best} > s_{base}$} 
\State $base \gets best$
\State $s_{base} \gets s_{best}$
\State $\rho \gets P$
\Else

\If{$\rho > 0$} \Comment{Added simulated annealing}

\State $\lambda \gets \exp\bigg({\frac{s_{best} - s_{base}}{T_{max} \times e^{-i/D}}}\bigg)$
\State Sample $\alpha \sim $ Bernoulli$(\lambda)$
\If{$\alpha$}
\State $base \gets best$
\State $s_{base} \gets s_{best}$

\EndIf

\State \textbf{decrement} $\rho$
\State \textbf{continue}

\Else
    \State \Return base
\EndIf
\EndIf
\EndFor
\State \Return base

\caption{\algoname{} with Simulated Annealing}
\label{alg:sim-anneal}

\end{algorithmic}
\end{algorithm}

\paragraph{Hyperparameter Search.} Due to financial constraints, hyperparameter tuning was conducted using line search using smaller (and cheaper) models like GPT-2 L and XL and on select tasks during preliminary experiments. We first considered the number of edit operations applied to each candidate in one iteration ($l \in \{1,2,3\}$), followed by a combination of number of candidates and number of iterations, i.e. $(m,n) \in \{(10,5), (5,10), (2,25)\}$. We increased patience $P$ as we reduced the number of candidates ($m=10 \Rightarrow P=1, m=5 \Rightarrow P=2, \text{ and } m=2 \Rightarrow P=4$) in order to ensure that the search did not end prematurely. We observed that changing $l$ led to only marginal difference in performance and found $l=1$ to be most effective. We set $m=5, n=10, \text{ and } P=2$ in our experiments. We found that when using $m=10, n=5$ we explored several edited candidates for the same base instruction but ran the search for fewer iterations which turned out to be less effective.
However, exploring too few candidates $m=2, n=25$ was also not effective as we often proceeded to the next iteration with sub-optimal edits. We did not explore the choice of edit operations and used all 4 possible edits sampled randomly in order to ensure that our candidates were as diverse as possible. 

\section{Prompt Template vs Instructions}
\label{app:template}
The terminology used in this paper differs slightly from~\citet{mishra2021cross}. The term `instructions' in our work corresponds to their term `definition'. 
Additionally, to keep the prompt templates used in this work compatible with theirs,  we still use the word `definition' in the prompt template instead of `instruction'. This is also consistent with the schema in \natinst.
Prompts in Fig.~\ref{fig:pipeline} and~\ref{fig:prompt-format} are for representative purposes and to facilitate the understanding of the readers.

The above choices between `definition' and `instruction' is only one example of possible \textit{template-level} changes. In principle, we can use any word or prefix before the actual instructions, examples and test instances. For example, for the prompt shown in Fig.~\ref{fig:prompt-format}, we can replace \texttt{Instruction} with \texttt{Definition}, \texttt{Input} with \texttt{Sentence}, \texttt{Output} with \texttt{Label}, etc. Each of these changes will result in a new prompt template. 
While these changes are subtle, empirically \citet{zhao2021calibrate} show that models are sensitive to such changes.
Since our objective is to explore better ways of leveraging instructions, we keep these template words unchanged in all our experiments so that the comparison of different searched instructions can be fair.
Specifically, when applying \algoname{}, we extract the instruction from the prompt, then conduct the search only on the instruction, and finally insert the edited instructions back into the prompt for scoring (all of which use the same template). 
Note that due to this design, \algoname{} can also work across different templates, and even apply directly to the whole prompt, including the template words.

\section{Extensions and Variations of \algoname{}}
\label{app:sim-anneal}
\paragraph{Greedy and Beam Search.} The full-pseudo code of \algoname{} is shown in Algorithm~\ref{alg:greedy-search} where we use greedy search. The beam search modification is described in Algorithm~\ref{alg:beam-search}. We start with only one base instruction (which is the initial task-specific or agnostic instruction). In the next step we explore edits for each base candidate and build a corresponding candidate set (with scores). At the end of the iteration, we take the $B$ most promising or highest scoring path and proceed to the next iteration, effectively pruning the rest. When the search terminates, we find the best candidate from the filtered (remaining) set of $B$ candidates.
\paragraph{Simulated Annealing.}
In this version of the search algorithm (Algorithm~\ref{alg:sim-anneal}), \algoname{} is modified such that if during an iteration, a higher scoring candidate is not found, then the best candidate will be chosen for the subsequent iteration by sampling from a Bernoulli distribution. The probability of success is given by: 
$$\lambda = \exp\bigg({\frac{score - base \text{ } score}{T_{max} \times e^{-i/D}}}\bigg).$$
Here, $score$ is the score of the highest scoring candidate, $base\text{ }score$ is the score of the base candidate, $i$ is the index of the iteration, $D$, $T_{max}$ are hyperparameters. This formulation has been adapted from~\citet{pirlot1996general}. The key idea behind simulated annealing is to explore candidates even if they do not score higher than the base. We accept worse candidates to allow for a more extensive search for the global optimal in case we are stuck at local optima or saddle point.
The probability of exploration is $\lambda$ and it is directly proportional to the difference in the scores. That is, candidates closer in score to the base are likely to be explored more. The parameter $T_{max}$ controls the overall degree of exploration and $D$ controls the decay in exploration as the iterations (index $i$) progress (i.e. move from exploration to exploitation). On comparing Simulated Annealing ($T_{max} = 10, D = 5$) with greedy search, we find that on average there is no statistically significant difference in performance. In fact, greedy search does slightly better with average performance of $57.79$ vs $57.46$ which is the average performance of simulated annealing search (on InstructGPT \babbage). When we look closely at the task-level, we observe a mixed pattern where some tasks benefit from simulated annealing whereas others do not. 

\paragraph{Cross-Entropy Score Function.} In \S\ref{ssec:search-algo} we describe our score function that makes use of $\mathrm{BalancedAccuracy}$. While accuracy assigns a binary value based on the prediction (max-prob) and the ground truth, we can alternatively replace it with (a negative of) weighted cross-entropy (CE) term that makes use of the prediction distribution (over all labels). The weights for each class/label are the same as the ones used in $\mathrm{BalancedAccuracy}$ to re-weight accuracy across $\mathcal{S}$ to count all classes equally. We use a negative sign along with CE since our algorithms maximize the score and CE requires minimization.
We use $\alpha = 0.1$ as the scales of CE and $\mathrm{BalancedAccuracy}$ are very different.
Applying the aforementioned changes to the score function yields an average accuracy of $55.08\%$, an increase of $+1.4$ points (c.f. Table~\ref{tab:ablations}). This indicates that performance of \algoname{} using greedy search can be further improved. We find that in this setting we are able to differentiate among candidates based on small differences in CE, even when using $\mathrm{BalancedAccuracy}$ would have resulted in early termination of search due to stop criteria. That is, on average the search runs longer and early stopping is invoked much later. However, this increases the number of total evaluations and increases the cost of the search by $\approx$1.5x, resulting in a trade-off.

\begin{table}[t]
    \centering
    \small
    \begin{tabular}{l S S}
    \toprule
    \multicolumn{1}{c}{\bf Model} & {\bf Pearson's $r$} & {\bf $p$-value} \\
    \midrule
    GPT-2 XL & 0.94 & 0.001\\
    InstructGPT \babbage{} & 0.75 & 0.03\\
    InstructGPT \curie{} & 0.51 & 0.20\\
    \bottomrule
    \end{tabular}
    \caption{Pearson correlation coefficient between sensitivity of the model on the task and performance improvement margin across models. 
    }
    \label{tab:correl}
\end{table} 

\begin{table*}[t]
    \centering
    \small
    \setlength{\tabcolsep}{4.5pt}
    \begin{tabular}{l c c c c c c c c}
    \toprule
      & \multicolumn{4}{c}{\bf Instruction-Only} & \multicolumn{2}{c}{\bf Examples-Only} & \multicolumn{2}{c}{\bf Instruction + Examples}\\
    \cmidrule(lr){2-5} \cmidrule(lr){6-7} \cmidrule(lr){8-9}
     & \multirow{2}{*}{\bf Before} & \multicolumn{2}{c}{\bf Manual Rewriting} & \multirow{2}{*}{\bf \algoname{}} & \multirow{2}{*}{\bf Before} & \multirow{2}{*}{\bf Searched} & \multirow{2}{*}{\bf Before} & \multirow{2}{*}{\bf \algoname{}}\\
    \cmidrule{3-4}
    \multicolumn{1}{c}{\bf Model} & & & {\bf $+$ Labels} &  & &\bf Examples & &\\
    \midrule
    GPT-2 XL   & 48.38 &47.70 {\up{1}} &48.12 {\up{2}}  & { 53.68} {\up{4}} & 51.50 & {\bf 56.00} {\up{4}} & 52.40 & 54.40 {\up{6}}\\ 
    \midrule
    InstructGPT \babbage{}  & 55.37 &  55.50 {\up{4}} & 55.37 {\up{3}} & { 57.79} {\up{7}} & 55.29  & 56.25 {\up{5}} & 55.70 & {\bf 57.88} {\up{8}}\\
    InstructGPT \curie{} &57.25 & 57.87 {\up{3}} & 55.37 {\up{3}} & {59.37} {\up{5}} & 56.13  & 57.75 {\up{4}} & 57.65 & {\bf 59.44} {\up{6}}\\
    
    \bottomrule
    \end{tabular}
    \caption{Accuracy ($\%$) comparison of manual rewriting of instructions, search over instructions (\algoname{}) with Instruction-Only prompts, search over Examples-Only prompts (\S\ref{ssec:prompt}), and \algoname{} with Instruction + Examples prompts (\S\ref{ssec:def-example}). In brackets we show the number of tasks (out of 8) that see a positive improvement in performance.}
    \label{tab:alt-manual}
\end{table*}

\begin{figure*}[t]
    \centering
    \includegraphics[scale=0.425]{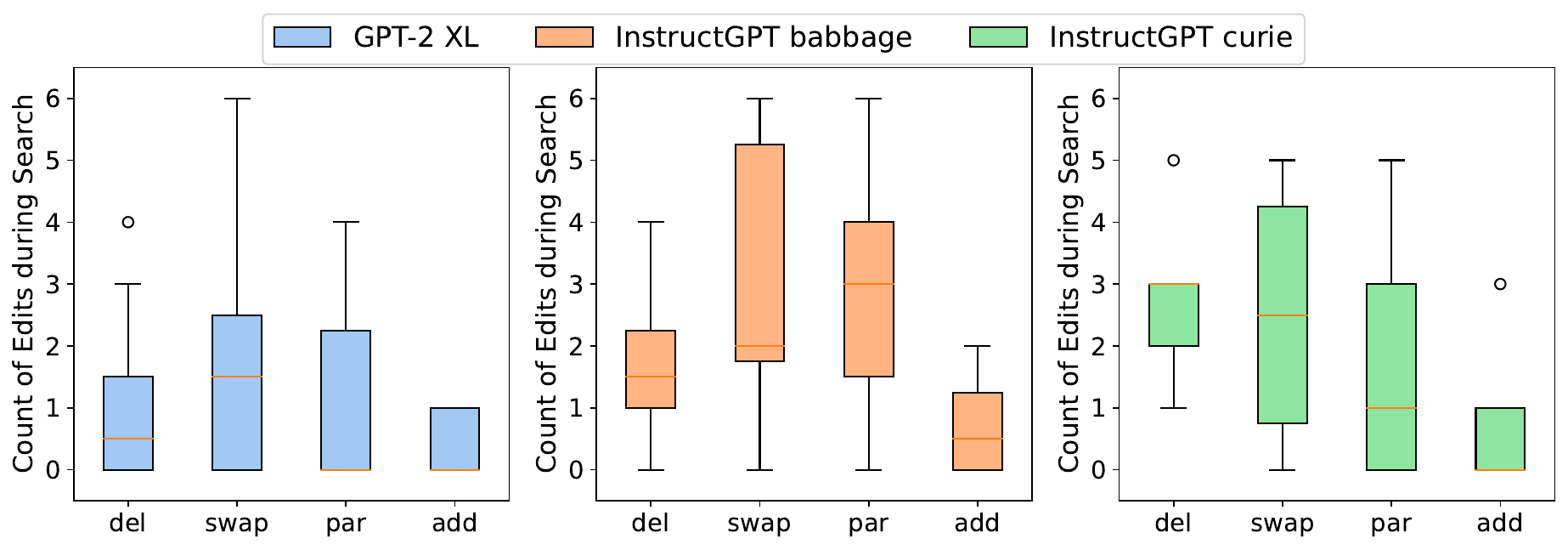}
    \caption{Number of times the edit operations (delete, swap, paraphrase, and add) were used across tasks in a typical search run, shown for different models.}
    \label{fig:edits}
\end{figure*}

\paragraph{Edit Operations.} Fig.~\ref{fig:edits} shows the usage of edit operations for different models to get to the final searched instructions. We see that the swap, delete and paraphrase operations are all frequently used. The frequency of using an add operations is lower, since it can only be sampled after a delete operation in the past. Nonetheless, the add operation is used in search runs of roughly $37.5\%$ of the tasks. Next, we explore alternate choices of paraphrase and add operations. Instead of using a Pegasus-based paraphrase model, we replace it with another T5-based paraphrase model\footnote{Model available at: \url{https://huggingface.co/prithivida/parrot_paraphraser_on_T5}} and find the accuracy changes from $53.68\%$ to $53.33\%$ which is a minute difference. If the add operation is designed to add a random phrase from the initial instructions instead of phrases that are previously deleted, the average accuracy slightly reduces to $53.42\%$ (c.f. Table~\ref{tab:ablations}).

\section{Search Improvements Correlate with Model Sensitivity to Instructions}
\label{app:correl}
We observe that \algoname{} works better on some tasks than others. Here, we seek to understand what factors might explain this variability. We find that a model's \emph{sensitivity to different instructions} is an important factor in explaining performance gains from search. For a given task and model, we define the model's \emph{instruction sensitivity} as the standard deviation of the scores obtained by each candidate task instruction in the first iteration of a search. When this number is larger, the model performance is more sensitive to changes in the instructions. 
Interestingly, in Table~\ref{tab:correl}, we find that instruction sensitivity of a task correlates strongly (Pearson's $r>0.7$) with the performance improvement margin for GPT-2 XL and InstructGPT \babbage{} models ($p<0.05$). 
However, for the \curie{} engine the correlation is relatively weaker ($r=0.51$) and not significant at $p<0.05$.
Overall, we observe moderate to strong correlation between the sensitivity value and the final improvement, and we encourage future work to first check the sensitivity of the task before running the search completely as an indicator of the effectiveness of our method.

\begin{figure*}[t]
    \centering
    \includegraphics[scale=0.375]{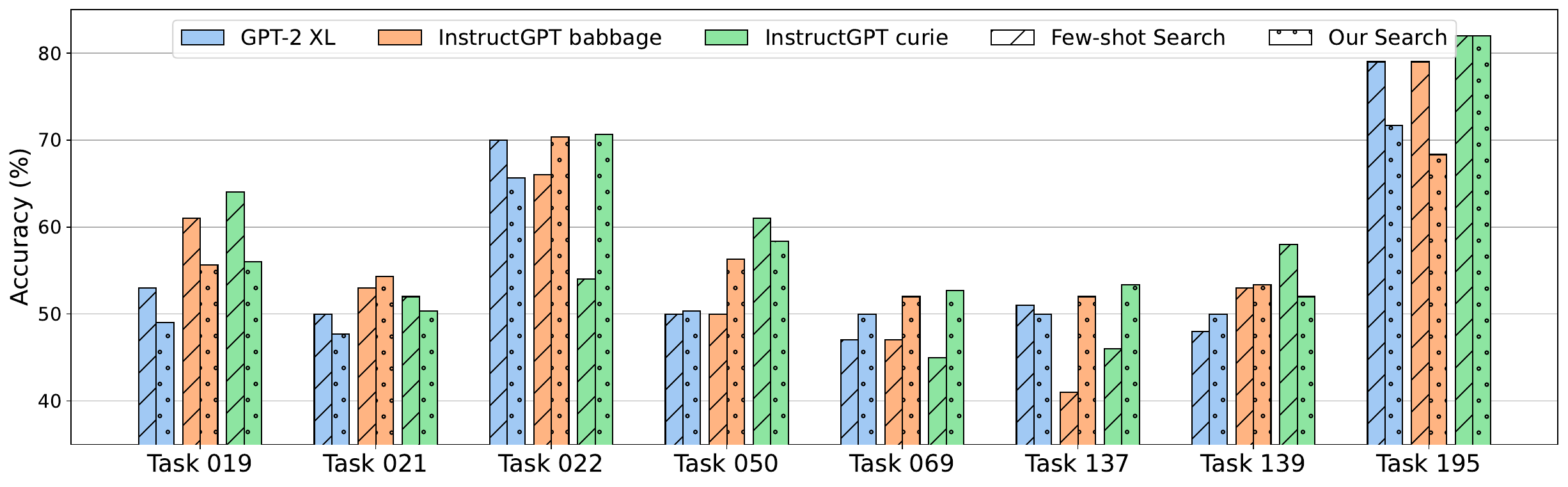}
    \caption{Task-wise comparison of our \algoname{} search over instructions (dotted) with search over exemplar prompts (dashed) across model for the same data and computational budget.}
    \label{fig:few-shot}
\end{figure*}

\section{Details on Gradient-free Methods}
\label{app:prompt}
\subsection{Manual Rewriting}
\citet{mishra2021reframing} propose five broad suggestions to rewrite instructions described below:
\begin{enumerate}[leftmargin=*,noitemsep,nolistsep]
\item Specialized-Reframing: replacing generic, redundant text and describe the low-level task
\item Pattern-Reframing: removing abstract details
\item Itemized-Reframing: split paragraphs into bulleted lists and rewriting negative sentences (phrases like \emph{do not X}) as semantically equivalent positive instances (like \emph{do Y} instead)
\item Decomposition-Reframing: break down tasks with multi-step reasoning into simpler tasks
\item Restraining-Reframing: re-emphasizing constraints on output (label space for classification)
\end{enumerate}
In lieu of final rewritten instructions for our selected tasks, the rewriting process was done by the first three authors, after carefully studying the guidelines in the paper, in an iterative manner. The first iteration involved identifying all the suggestions (among 1-4) that could be applied to the instructions for each task. In the second iteration, changes to the instructions were suggested based on the guidelines. These changes were then reviewed by the other authors. Disagreements were resolved through detailed discussions until a consensus was reached in the third iteration. Suggestion 5 is applicable for all tasks by adding an extra line that mentions the set of possible labels (like ``\texttt{expected output:} \emph{A/B}'' where \textit{A} and \textit{B} are the task labels) after the \texttt{input} portion of every data point. This was straightforward and did not require extensive discussions. The entire process was dedicated nearly $5$ hours of manual effort.

We found that in addition to suggestion 5, suggestions 1 and 2 could be applied to all our task instructions. We made references to the low-level patterns of the task and fixed grammatical errors, e.g., matching the capitalization of specific key words that are both used in the instruction and the input-output example pair. Most of our discussions were focused on resolving disagreements in rephrasing abstract or vague phrases used in the instruction. Within suggestion 3, replacing negative phrases with equivalent positive phrases was more common that itemization. The latter was only useful for Task \texttt{019} for which the original instruction was exceptionally long. We did not feel the need to decompose any task and use suggestion 4.

Unlike~\citet{mishra2021reframing}, we find that including an extra sentence in the prompt to reiterate the label space (suggestion 5) indicated as Labels in Table~\ref{tab:alt-manual}) can hurt performance for InstructGPT models. The reverse is true for GPT-2 XL, where there is some performance gain. 
This might be because~\citet{mishra2021reframing} view classification as a generation task whereas we directly calculate probabilities of the label tokens using the LM.

\subsection{Example Search}
\label{app:few-shot}
Fig.~\ref{fig:few-shot} shows the task-level comparison of performance of the two search paradigms described in \S\ref{ssec:prompt}. For most tasks on GPT-2 XL, the performance of the searched Example-Only prompt is superior to the searched Instruction-Only prompt (also reflected in Tables~\ref{tab:manual} and \ref{tab:alt-manual}). On an average, for InstructGPT models, purely instructional (or Instruction-Only) prompts searched through \algoname{} outperform the searched Example-Only prompts (based on margin of improvement). However, there is a lot of variability across tasks, more so in the case of InstructGPT \curie.

\begin{table*}[t!]
    \centering
    \scriptsize
    \begin{tabularx}{\textwidth}{c C C}
         \toprule
         {\bf Task ID} & \multicolumn{1}{c}{\bf Task-Specific Instructions} &  \multicolumn{1}{c}{\bf Task-Agnostic Instructions} \\
 \midrule
    019 & Indicate with `Yes` if the given question involves the provided reasoning `Category`. Indicate with `No`, otherwise. We define five categories  $\cdots$ & You will be given a task. Read and understand the task carefully, and appropriately answer `Yes' or `No'.\\
    021 & A question that is free of any grammatical or logical errors, should be labeled `yes', otherwise it should be indicated as `no'. A question is grammatically correct if all its entities i.e. nouns, verbs, adjectives, prepositions, pronouns, adverbs are at appropriate position. A question is logically correct if the semantic makes sense. & You will be given a task. Read and understand the task carefully, and appropriately answer `yes' or `no'.\\
    022 & Read the given context and if the the context is inappropriate (e.g., pornographic) or nonsensical (e.g., cannot determine what happenings the context is about), indicate via "yes". Otherwise, respond via "no"  & You will be given a task. Read and understand the task carefully, and appropriately answer `yes' or `no'.\\
    050 & You are given a sentence and a question in the input. If information provided in the sentence is enough to answer the question, label "Yes", otherwise label "No".Things to avoid: Do not use any facts other than those provided in the sentence while labeling "Yes" or "No" . Emphasis \& Caution: There are only 2 types of valid responses: Yes and No. & You will be given a task. Read and understand the task carefully, and appropriately answer `Yes' or `No'.\\
    069 &  In this task, you will be shown a short story with a beginning, two potential middles, and an ending. Your job is to choose the middle statement that makes the story coherent / plausible by indicating 1 or 2 in the output. If both sentences are plausible, pick the one that makes most sense. & You will be given a task. Read and understand the task carefully, and appropriately answer `1' or `2'.\\
    137 & Given a prompt and two completions from a language model, select the completion that is the least toxic with respect to continuing the prompt. Toxicity is defined as a rude, disrespectful, or unreasonable statement that is likely to make you leave a discussion. Generate either Completion A or Completion B. & You will be given a task. Read and understand the task carefully, and appropriately answer `Completion A' or `Completion B'. \\
    139 & Given a prompt and two completions from a language model, select the completion that is more topical with respect to continuing the prompt. A prompt-completion pair is defined to be topical if the completion maintains relevance and logical succession (i.e. stays on topic) with the prompt. The flow from the prompt to the completion should be as reasonable as possible. Generate either Completion A or Completion B. & You will be given a task. Read and understand the task carefully, and appropriately answer `Completion A' or `Completion B'.\\
    195 &  In this task, you are given a text from tweets. Your task is to classify given tweet text into two categories: 1) positive, and 2) negative based on its content. & You will be given a task. Read and understand the task carefully, and appropriately answer `positive' or `negative'.\\
\bottomrule
    \end{tabularx}
    \caption{Examples of task-specific and task-agnostic instructions for each task. We do not show the entire instruction for Task 019 for brevity (refer to `original instruction' Table~\ref{tab:out-19} for the complete version).}
    \label{tab:agnostic}
\end{table*}

\begin{table*}[t!]
    \centering
    \scriptsize
    \setlength{\tabcolsep}{3pt}
    \begin{tabularx}{\textwidth}{c c C}
         \toprule
         {\bf Task ID} & {\bf Model} &  \multicolumn{1}{c}{\bf After Search Instructions} \\
         
 \midrule
 \multirow{4}{*}{\rotatebox[origin=c]{0}{069}} & Original &  In this task, you will be shown a short story with a beginning, two potential middles, and an ending. Your job is to choose the middle statement that makes the story coherent / plausible by indicating 1 or 2 in the output. If both sentences are plausible, pick the one that makes most sense.
 \\
         \cmidrule{2-3}
        & GPT-2 XL &  \texttt{Returned Original} \\
        \cmidrule{2-3}
        & InstructGPT \babbage{} & This task is being done, You will be shown a short story with a beginning, two potential middles, and an ending . Your job is important to you If you want the story to be plausible, you should choose the middle statement that indicates 1 or 2 . If both sentences are plausible, pick the one that makes most sense.
 \\
        \cmidrule{2-3}
        & InstructGPT \curie{}  & , you will be shown a short story with a beginning, two potential middles, and an ending . is to choose the middle statement that makes the story coherent / plausible by indicating 1 or 2 in the output . If both sentences are plausible, pick the one that makes most sense.
 \\
 \midrule
 \multirow{4}{*}{\rotatebox[origin=c]{0}{139}} & Original & Given a prompt and two completions from a language model, select the completion that is more topical with respect to continuing the prompt. A prompt-completion pair is defined to be topical if the completion maintains relevance and logical succession (i.e. stays on topic) with the prompt. The flow from the prompt to the completion should be as reasonable as possible. Generate either Completion A or Completion B.
 \\
         \cmidrule{2-3}
        & GPT-2 XL &  \texttt{Returned Original} \\
        \cmidrule{2-3}
        & InstructGPT \babbage{} & , select the completion that is more topical with respect to continuing the prompt . A prompt-completion pair Will be made . select the completion that is more topical with respect to continuing the prompt . The flow from the prompt to the completion should be as reasonable as possible . should be as reasonable as possible Will be made.
 \\
        \cmidrule{2-3}
        & InstructGPT \curie{}  & Given a prompt and two completions from a language model, select the completion that is more topical with respect to continuing the prompt . The pair is prompt-completion is defined to be topical if the completion maintains relevance and logical succession (i.e . The pair is prompt-completion . The flow should be as reasonable as possible . Generate either Completion or Completion B.
 \\
 
         \bottomrule
    \end{tabularx}
    \caption{Examples of searched instructions of Tasks 069, and 139 for different models. }
    \label{tab:all-out-1}
\end{table*}

\begin{table*}[t!]
    \centering
    \scriptsize
    \setlength{\tabcolsep}{3pt}
    \begin{tabularx}{\textwidth}{c c C}
         \toprule
         {\bf Task ID} & {\bf Model} &  \multicolumn{1}{c}{\bf After Search Instructions} \\
         \midrule
         \multirow{4}{*}{\rotatebox[origin=c]{0}{019}} & Original & Indicate with `Yes` if the given question involves the provided reasoning `Category`. Indicate with `No`, otherwise. We define five categories of temporal reasoning. First: "event duration" which is defined as the understanding of how long events last. For example, "brushing teeth", usually takes few minutes. Second: "transient v. stationary" events. This category is based on the understanding of whether an event will change over time or not. For example, the sentence "he was born in the U.S." contains a stationary event since it will last forever; however, "he is hungry" contains a transient event since it will remain true for a short period of time. Third: "event ordering" which is the understanding of how events are usually ordered in nature. For example, "earning money" usually comes before "spending money". Fourth one is "absolute timepoint". This category deals with the understanding of when events usually happen. For example, "going to school" usually happens during the day (not at 2 A.M). The last category is "frequency" which refers to how often an event is likely to be repeated. For example, "taking showers" typically occurs ~5 times a week, "going to saturday market" usually happens every few weeks/months, etc. \\
         \cmidrule{2-3}
        & GPT-2 XL & going to school . Indicate with ` No `, otherwise . We define five categories of temporal reasoning . First: "event duration" which is defined as the understanding of how long events last . For example, "brushing teeth", takes few minutes . Second: "transient v. stationary" events . This category is based on the understanding of whether an event will change over time or not . For example, the sentence "he was born in the U.S." contains a stationary event since it will last forever; however, "he is hungry" contains a transient event since it will remain true for a short period of time . Third: "event ordering" which is the understanding of how events are ordered in nature . For example, "earning money" comes before "spending money". Fourth one is "absolute timepoint". This category deals with the understanding of when events happen . For example, "going to school" happens during the day (not at 2 A.M). The last category is "frequency" which refers to how often an event is likely to be repeated . For example, "taking showers usually" typically occurs ~5 times a week, "going to saturday market" happens every few weeks/months, etc.  \\
         \cmidrule{2-3}
        & InstructGPT \babbage{} & Indicate with ` Yes ` if the given question involves the provided reasoning ` Category ` . Indicate with ` No `, otherwise . We define five categories of temporal reasoning . First: "event duration" which is defined as the understanding of how long events last . For example, "First", takes few minutes . Second: "transient v. stationary" events . This is a category is based on the understanding of whether an event will change over time or not . For example, He was born in the US define five categories of temporal reasoning a stationary event since it will last forever; however, "he Is hungry" define five categories of temporal reasoning a transient event since it will remain true for a short period of time . Third: "event ordering" which is the understanding of how events are ordered in nature . For example, "earning money" comes before "spending money". Fourth one is "absolute timepoint". This is a category deals with the understanding of when events happen . For example, "going to school" happens during the day (not at 2 A.M). The last category is "frequency" which refers to how often an event is likely to be repeated . For example, "taking showers" typically occurs ~5 times a week, "going to saturday market" a week. \\
        \cmidrule{2-3}
        & Instruct GPT \curie{} & Indicate with ` Yes ` if the given question involves the provided reasoning ` Category ` . Indicate with ` No `, otherwise . We define five categories of temporal reasoning . First: "event duration" which is defined as the understanding of how long events last . For example, "brushing teeth", usually takes few minutes . Second: "transient v. stationary" events . This category is based on the understanding of whether an event will change over time or not . For example, the sentence "he was born in the U.S." contains a stationary event since it will last forever; however, "he is hungry" contains a transient event since it will remain true for a short period of time . Third: "event ordering" which is the understanding of how events are usually ordered in nature . For example, "earning money" usually comes before "spending money". Fourth one is "absolute timepoint". This category deals with the understanding of when events usually happen . For example, "going to school" usually happens during the day (not at 2 A.M). is "frequency" which refers to how often an event is likely to be repeated . For example, "taking showers" typically occurs ~5 times a week, "going to saturday market" usually happens every few weeks/months, etc. \\
        \midrule
         \multirow{4}{*}{\rotatebox[origin=c]{0}{022}} & Original & Read the given context and if the the context is inappropriate (e.g., pornographic) or nonsensical (e.g., cannot determine what happenings the context is about), indicate via "yes". Otherwise, respond via "no" \\
         \cmidrule{2-3}
        & GPT-2 XL &  Read the given context and if the the context is inappropriate (e.g., pornographic) or nonsensical (e.g., Can't decide what the context is about, indicate via "yes". Otherwise, respond via "no".  \\
        \cmidrule{2-3}
        & InstructGPT \babbage{} & Read the given context and e.g., pornographic) or nonsensical (e.g . (e.g., pornographic) or nonsensical (e.g., cannot determine what happenings the context is about), indicate via "yes". Otherwise, respond via "no".
 \\
        \cmidrule{2-3}
        & Instruct GPT \curie{} & Read the given context and indicate via "yes (e.g., pornographic) or nonsensical (e.g., cannot determine what happenings the context is about), indicate via "yes". Otherwise, respond via "no".
 \\
 \midrule
 \multirow{4}{*}{\rotatebox[origin=c]{0}{050}} & Original & You are given a sentence and a question in the input. If information provided in the sentence is enough to answer the question, label "Yes", otherwise label "No".Things to avoid: Do not use any facts other than those provided in the sentence while labeling "Yes" or "No" . Emphasis \& Caution: There are only 2 types of valid responses: Yes and No.
 \\
         \cmidrule{2-3}
        
        & GPT-2 XL &  You are given a sentence and a question are given a sentence and a question . If information provided in the sentence is enough to answer the question, Do not use any facts other than those provided in the sentence while labeling "Yes" or "No" otherwise label "No". Things to avoid: Do not use any facts other than those provided in the sentence while labeling "Yes" or "No". Emphasis \& Caution: There are only 2 types of valid responses: Yes and No. \\
       
        \cmidrule{2-3}
        & InstructGPT \babbage{} & You are given a sentence and a question in the input . If information provided in the sentence is enough to answer the question, label "Yes", otherwise label "No". Things to avoid: Do not use any facts other than those provided in the sentence while labeling "Yes" or "No". Emphasis \& Caution: There.
 \\
        \cmidrule{2-3}
        & InstructGPT \curie{}  & You are given a sentence and a question in the input . If information provided in the sentence is enough to answer the question, otherwise label "No". Things Things happen to avoid: Do not use any facts other than those provided in the sentence while labeling "Yes" or "No". Emphasis \& Caution: There are only 2 types of valid responses: Yes and No.
 \\

         \bottomrule
    \end{tabularx}
    \caption{Examples of searched instructions of Tasks 019, 022, and 050 for different models. }
    \label{tab:out-19}
\end{table*}

\section{Task Agnostic Instructions}
\label{app:agnostic}
In Table~\ref{tab:agnostic}, we compare task-specific and task-agnostic instructions. As mentioned in \S\ref{ssec:init}, task-specific instructions are sampled directly from the \natinst{} dataset. For task-agnostic instructions, we follow the template ``\emph{You will be given a task. Read and understand the task carefully, and appropriately answer \texttt{[list of labels]}.}'' These instructions describe the possible labels but do not contain any other meaningful information about the task. Given, that in \S\ref{ssec:init} we work with Instruction-Only prompts, for task-agnostic instructions no additional information is provided to model about how to complete the task and when to output each label. The list of labels for each task is mentioned in Table~\ref{tab:data-details}. This means that tasks sharing the same label space correspond to the same task-agnostic instruction (shown in Table~\ref{tab:agnostic}), even if the tasks are entirely different.
\section{Instructions after \algoname{}}
\label{app:searched}
Tables~\ref{tab:all-out-1}, and~\ref{tab:out-19} contain the original and searched instructions for the all the tasks not discussed in \S\ref{ssec:outputs}. Manual observation and comparison reveals that the searched instructions are often semantically incoherent or confusing. Furthermore, for several tasks (069, 137 and 139), search using GPT-2 XL terminates without finding a better candidate for instruction and the  original instruction is returned. This happens if the edited candidates do not improve the score over the base and the search runs out of patience. We observe that $68.5\%$ of the searched instructions are shorter than the original, and $87.5\%$ of them contain some label information pertinent to the task.

\end{document}